\documentclass[10pt,twocolumn,letterpaper]{article}

\usepackage{cvpr}
\usepackage{times}
\usepackage{epsfig}
\usepackage{graphicx}
\usepackage{amsmath}
\usepackage{amssymb}
\usepackage{bm}
\usepackage{subfig}
\usepackage[font={small}]{caption}

\usepackage[breaklinks=true,bookmarks=false]{hyperref}
\hypersetup{hidelinks}

\cvprfinalcopy 

\def\cvprPaperID{****} 

\ifcvprfinal\fi
\begin{document}

\title{Mask-Guided Portrait Editing with Conditional GANs}

\author{Shuyang Gu$^{1}$ \qquad Jianmin Bao$^{1}$ \qquad Hao Yang$^{2}$ \qquad Dong Chen$^{2}$ \qquad Fang Wen$^{2}$ \qquad Lu Yuan$^{2}$ \vspace{1pt}\\
$^{1}$University of Science and Technology of China  \qquad $^{2}$Microsoft Research\qquad\qquad\\
\hspace{0.1in}{\tt\small \{gsy777,jmbao\}@mail.ustc.edu.cn} \qquad  {\tt\small \{haya,doch,fangwen,luyuan\}@microsoft.com} \\
}

\twocolumn[{%
\renewcommand\twocolumn[1][]{#1}%
\vspace{-1em}
\maketitle
\vspace{-1em}
\begin{center}
    \centering
    \vspace{-0.3in}
    \includegraphics[width=\linewidth]{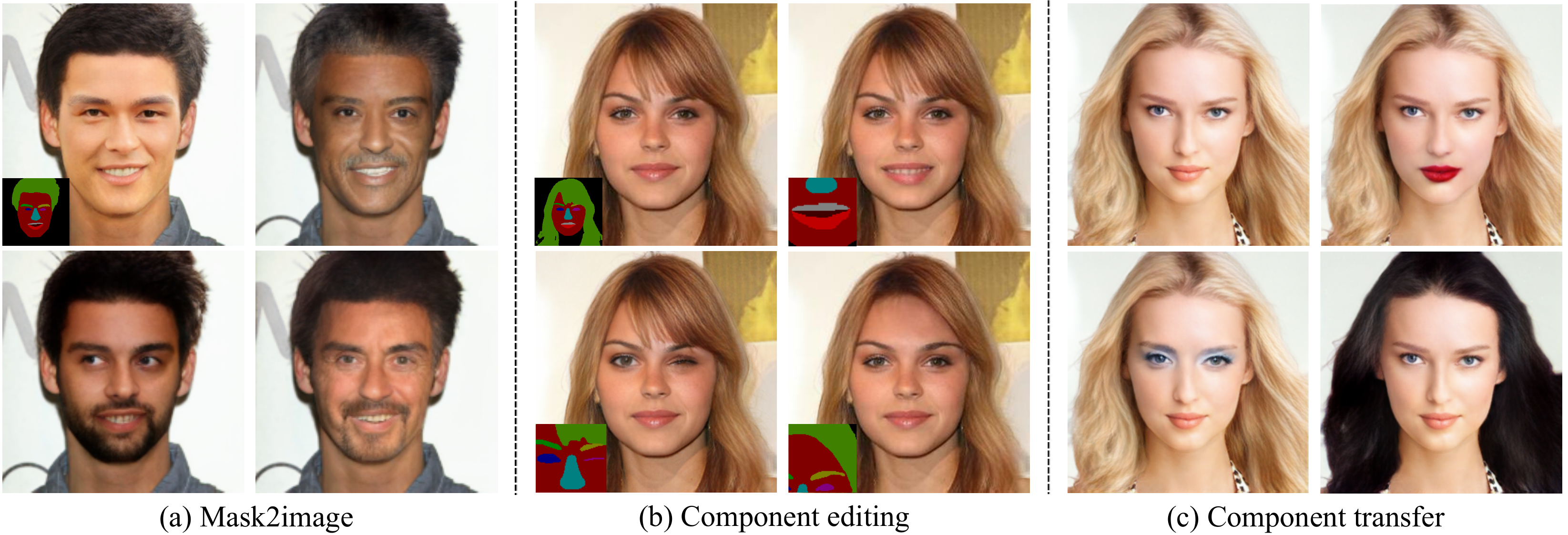}
    \vspace{-0.25in}
    \captionof{figure}{We propose a framework based on conditional GANs for mask-guided portrait editing. (a) Our framework can generate diverse and realistic faces using one input target mask (lower left corner in the first image). (b) Our framework allows us to edit the mask to change the shape of face components, \ie mouth, eyes, hair. (c) Our framework also allows us to transfer the  appearance of each component for a portrait, including hair color. }
    \label{fig:teaser}
\end{center}%
}]

\maketitle
\thispagestyle{empty}
\vspace{-1.5em}
\begin{abstract}
Portrait editing is a popular subject in photo manipulation. The Generative Adversarial Network (GAN) advances the generating of realistic faces and allows more face editing. In this paper, we argue about three issues in existing techniques: diversity, quality, and controllability for portrait synthesis and editing. To address these issues, we propose a novel end-to-end learning framework that leverages conditional GANs guided by provided face masks for generating faces. The framework learns feature embeddings for every face component (\eg, mouth, hair, eye), separately, contributing to better correspondences for image translation, and local face editing. With the mask, our network is available to many applications, like face synthesis driven by mask, face Swap+ (including hair in swapping), and local manipulation. It can also boost the performance of face parsing a bit as an option of data augmentation.
\end{abstract}

\section{Introduction}

Portrait editing is of great interest in the vision and graphics community due to its potential applications in movies, gaming, photo manipulation and sharing, etc. People enjoy the magic that makes faces look more interesting, funny, and beautiful, which appear in an amount of popular apps, such as Snapchat, Facetune, etc.
 
Recently, advances in Generative Adversarial Networks (GANs)~\cite{goodfellow2014generative} have made tremendous progress in synthesizing realistic faces~\cite{antipov2017face,li2016convolutional,karras2017progressive,choi2017stargan}, like face aging~\cite{yang2017learning}, pose changing~\cite{tran2017disentangled,huang2017beyond} and attribute modifying~\cite{bao2018towards}. However, these existing approaches still suffer from some quality issues, like lack of fine details in skin, difficulty in dealing with hair and background blurring. Such artifacts cause generated faces to look unrealistic. 

To address these issues, one possible solution is to use the facial mask to guide generation. On one hand, a face mask provides a good geometric constraint, which helps synthesize realistic faces. On the other hand, an accurate contour for each facial component (\eg, eye, mouth, hair, etc.) is necessary for local editing. Based on the face mask, some works~\cite{shih2014style,fivser2017example} achieve very promising results in portrait stylization. However, these methods focus on transferring the visual style (\eg, B\&W, color, painting) from the reference face to the target face. It seems to be unavailable for synthesizing different faces, or changing face components.

Some kinds of GAN models begin to integrate the face mask/skeleton for better image-to-image translation, for example, pix2pix~\cite{isola2017image}, pix2pixHD~\cite{wang2017high}, where the facial skeleton plays an important role in producing realistic faces and enabling further editing. However, the diversity of their synthesized faces are so limited, for example, the input and output pairs might not allow noticeable changes in emotion. The quality issue especially on hair and background prevents resulting images from being realistic. The recent work BicycleGAN~\cite{zhu2017toward} tries to generate diverse faces from one input mask, but the diversity is limited on color or illumination.

We have to reinvestigate and carefully design the image-to-image translation model, which addresses the three problems -- \emph{diversity}, \emph{quality}, and \emph{controllability}. Diversity requires the learning of good correspondences between image pairs, which may undergo variance in poses, lightings, colors, ages and genders, for image translation. Quality should further improve in fine facial details, hair, and background. More controls for local facial components are also key.

In this paper, we propose a framework based on conditional GANs~\cite{mirza2014conditional} for portrait editing guided by face masks. The framework consists of three major components: \emph{local embedding sub-network}, \emph{mask-guided generative sub-network}, and \emph{background fusion sub-network}. The three sub-networks are trained end-to-end. The \emph{local embedding sub-network} involves five auto-encoder networks which respectively encode embedding information for five facial components, \ie, ``left eye", ``right eye", ``mouth", ``skin \& nose", and ``hair". The face mask is used to help specify the region for each component in learning. The \emph{mask-guided generative sub-network} recombines the pieces of local embeddings and the target face mask together, yielding the foreground face image. The face mask helps establish correspondences at the component level (\eg, mouth-to-mouth, hair-to-hair, etc.) between the source and target images. At the end, the \emph{background fusing sub-network} fuses the background and the foreground face to generate a natural face image, according to the target face mask. For guidance, the face mask aids facial generation in all of three sub-networks.

With the mask, our framework allows many applications. As shown in Figure~\ref{fig:teaser} (a), we can generate new faces driven by the face mask, \ie, mask-to-face, as well as skeleton-to-face in~\cite{isola2017image,wang2017high}. We also allow more editing, such as removing hairs, amplifying or reducing eyes, and making it smile, as shown in Figure~\ref{fig:teaser} (b). Moreover, we can modify the appearance of existing faces locally, such as the changing appearance of each facial component, shown in Figure \ref{fig:teaser} (c). Experiments shows that our methods outperform state-of-the-art face synthesis driven by a mask (or skeleton) in terms of diversity and quality. More interesting, our framework can help boost the performance of face parsing algorithm marginally as the data augmentation.

Overall, our contributions are as follows: \vspace{-0.5em}

 \begin{enumerate}
 \item We propose a novel framework based on mask-guided conditional GANs, which successfully addresses diversity, quality and controllability issues in face synthesis. \vspace{-0.5em}

 \item The framework is general and available for an amount of applications, such as mask-to-face synthesis, face editing, face swap+, and even data augmentation for face parsing. 
 \end{enumerate}





\section{Related Work}


\paragraph{Generative Adversarial Networks}

Generative adversarial networks (GANs) \cite{goodfellow2014generative} have achieved impressive results in many directions.
 It forces the generated samples to be indistinguishable from the target distribution by introducing an adversarial discriminator.
The GAN family enables a wide variety of computer vision applications such as image synthesis \cite{arjovsky2017wasserstein,qi2017loss}, image translation \cite{isola2017image,zhu2017unpaired,wang2017high,zhu2017toward}, and representation disentangling \cite{huang2018multimodal,bao2018towards,tran2017disentangled}, among others. 

Inspired by the conditional GAN models \cite{mirza2014conditional} that generate images from masks, we propose a novel framework for mask-guided portrait editing. Our method leverages local embedding information for individual facial components, generating portrait images with higher diversity and controllability, than existing global-based approaches such as pix2pix \cite{isola2017image} or pix2pixHD \cite{wang2017high}.

\vspace{-0.7em}

\paragraph{Deep Visual Manipulation}
Image editing has benefited a lot from the rapid growth of deep neural networks, including image completion \cite{iizuka2017globally}, super-resolution \cite{ledig2017photo}, deep analogy \cite{liao2017visual}, and sketch-based portrait editing \cite{sangkloy2017scribbler,portenier2018faceshop}, to name a few. 
Among them, the most related are mask-guided image editing methods, which train deep neural networks to translate masks into realistic images \cite{chen2017photographic,champandard2016semantic,isola2017image,wang2017high,zhu2017unpaired}.  

Our approach also relates to the visual attribute transfer methods, including style transfer \cite{gatys2016image,gu2018arbitrary} and color transfer \cite{he2017neural}. Recently, the Paired-CycleGAN \cite{chang2018pairedcyclegan} has been proposed for makeup transfer, in which a makeup transfer function and a makeup removal function are trained in pair. 
Though similar, the appearances of facial instances that our method disentangles differ from makeups. For example, the color and curly types of hairs which we can transfer are definitely not makeups. Furthermore, there are some works focusing on editing a specific component in faces (\eg, eyes \cite{dolhansky2018eye,shu2017eyeopener}) or editing attributes of faces \cite{pumarola2018ganimation, shu2017neural}. 

With the proposed structure that disentangles and recombines facial instance embeddings with face masks, our method also enhances over face swapping methods \cite{bitouk2008face,korshunova2017fast} by supporting explicit face and hair swapping.

\vspace{-0.7em}

\paragraph{Non-Parametric Visual Manipulation}

Non-parametric image synthesis approaches \cite{hays2007scene,busto2010instant,lalonde2007photo} usually generate new images by warping and stitching together existing patches from a database. The idea is extended by Qi \etal \cite{qi2018semi} which combines neural networks to improve quality.
Though similar at first glance, our method is intrinsically different from non-parametric image synthesis: our local embedding sub-network encodes facial instances as embeddings instead of image patches. New face images in our method are generated through the mask-guided generative sub-network, instead of warping and stitching image patches together. By jointly training all sub-networks, our model generates facial images that are higher quality than non-parametric methods that may also be difficult for them: \eg synthesizing a face with an open mouth showing teeth from a source face with a closed mouth hiding all teeth.

\begin{figure*}[t]
\centering
 \includegraphics[width=1.0\linewidth]{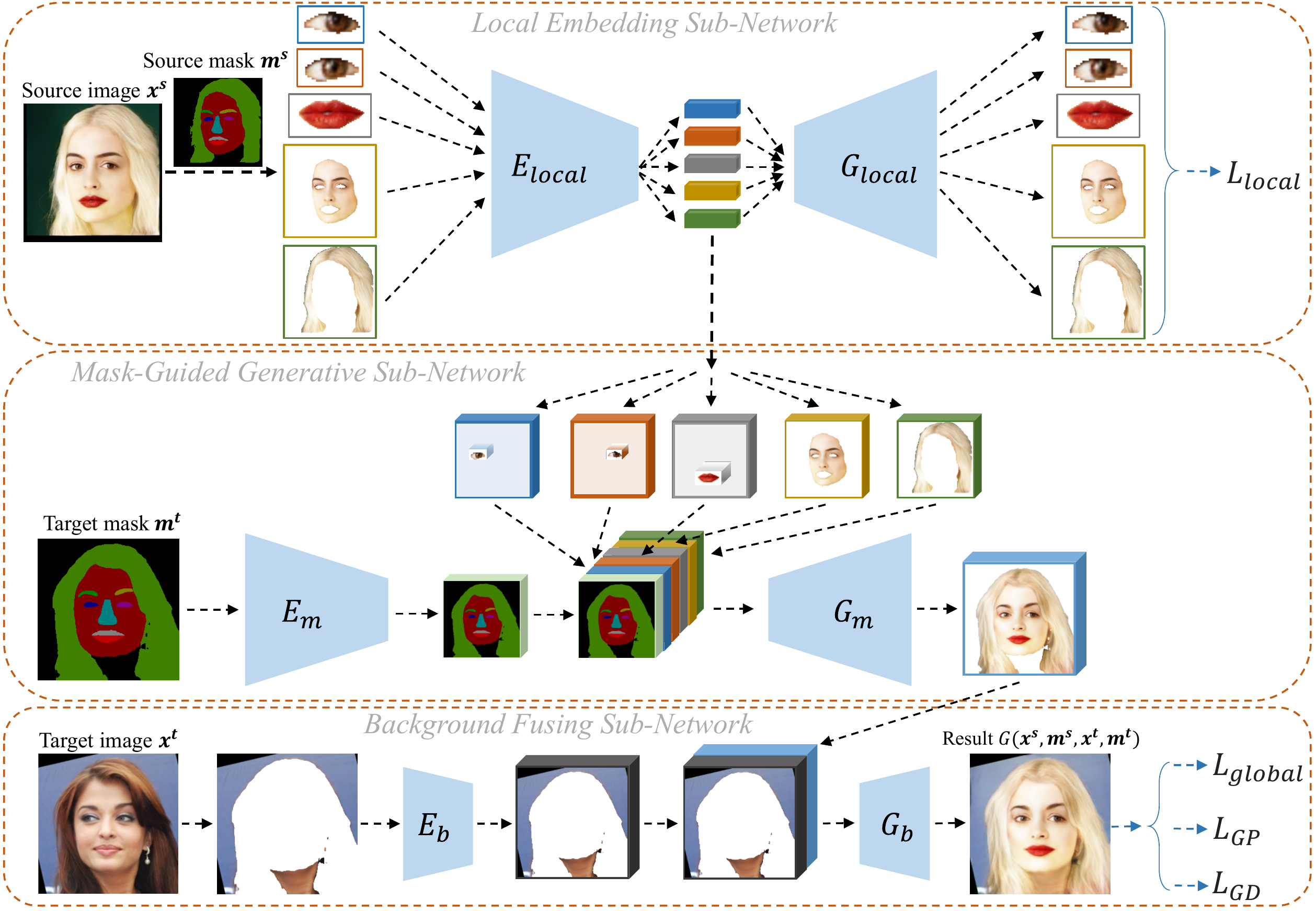}
 \vspace{-2.0em}
 \footnotesize
    \caption{The proposed framework for mask-guided portrait editing. It contains three parts: \emph{local embedding sub-network}, \emph{mask guided generative sub-network}, and \emph{background fusing sub-network}. \emph{Local embedding sub-network} learns the feature embedding of the local components of the source image. \emph{Mask guided sub-network} combines the learned component feature embeddings and mask to generate the foreground face image. \emph{Background fusing sub-network} generates the final result from the foreground face and the background. The loss functions are drawn with the blue dashed lines.}
\label{fig:framework}
\vspace{-2.0em}
\end{figure*}


\section{Mask-Guided Portrait Editing Framework}
We propose a framework based on conditional GANs for mask-guided portrait editing. Our framework requires four inputs, a source image ${\bm{x^s}}$, the mask of source image $\bm{m^s}$, a target image ${\bm{x^t}}$, and the mask of target image $\bm{m^t}$. The mask $\bm{m^s}$ and $\bm{m^t}$ can be obtained by a face parsing network. If we want to change the mask, we can manually edit $\bm{m^t}$. With the source mask $\bm{x^s}$, we cannot get the appearance of each face component, \eg, ``left eye", ``right eye", ``mouth", ``skin \& nose", and ``hair" from the source image. With the target mask $\bm{m^t}$, we can get the background from the target image $\bm{x^t}$. Our framework first recombines the appearance of each component from $\bm{x^s}$ and the target mask together, yielding the foreground face, then fusing it with the background from $\bm{x^t}$, outputting the final result $G(\bm{x^s}, \bm{m^s}, \bm{x^t}, \bm{m^t})$. $G$ indicates the overall generative framework.

As shown in Figure~\ref{fig:framework}, we first use a \emph{local embedding sub-network} to learn feature embedding for the input source image $\bm{x^s}$. It involves five auto-encoder networks to encode embedding information for five facial components respectively. Comparing with~\cite{wang2017high,zhu2017toward} which learn global embedding information, our approach can retain more source facial details. The \emph{mask guided generative sub-network} then specifies the region of each embedded component feature and concatenates all features of the five local components together with the target mask to generate the foreground face. Finally, we use the \emph{background fusing sub-network} to fuse the foreground face  and the background to generate a natural facial image.


\subsection{Framework Architecture}
\label{sec:Framework Architecture}

\paragraph{Local Embedding Sub-Network.}
To enable component-level controllability of our framework, we propose learning feature embeddings for each component in the face. We first use a face parsing network $P_F$ (details in Section~\ref{sec:loss functions}) which is a Fully Convolution Network (FCN) trained on the Helen dataset~\cite{smith2013exemplar} to get the source mask $\bm{m^s}$ of the source image $\bm{x^s}$. Then, according to the face mask, we segment the foreground face image into five components $\bm{x_i^s}$ , $i \in \{0, 1, 2, 3, 4\}$, \ie, ``left eye", ``right eye", ``mouth", ``skin \& nose", and ``hair". For each facial component, we use the corresponding auto-encoder network \{$E_{local}^i$, $G_{local}^i$\}, $i \in \{0, 1, 2, 3, 4\}$, to embed its component information. With five auto-encoder networks, we can conveniently change any one of facial components in the generated face images or recombine different components from different faces. 

Previous works, \eg, pix2pixHD~\cite{wang2017high}, also train an auto-encoder network to get the feature vector that corresponds with each instance in the image. To guarantee the features that fit different instance shape, they add an instance-wise average pooling layer to the output of the encoder to compute the average feature for the object instance. Although this approach allows object-level control on the generated results, their generated faces still suffer from low quality for two reasons. First, they use a global encoder network to learn feature embeddings for different instances in the image. We argue that merely a global network is quite limited in learning and recovering all local details of each instance. Second, the instance-wise average pooling would remove many characteristic details in reconstruction.
\vspace{-1.0em}


\paragraph{Mask-Guided Generative Sub-Network.}
\label{paragraph:Mask-Guided Generative Networks}
To make the target mask $\bm{m^t}$ a guidance for mask equivariant facial generation, we adopt an intuitive way to fuse the five component feature tensors and the mask feature tensor together. As shown in Figure~\ref{fig:framework} (b), five component feature tensors are extracted by the \emph{local embedding sub-network}, and the mask feature tensor is the output of the encoder $E_{m}$. 

First, we get the center location $\{c_i\}_{i=1\dots 5}$ of each component from the target mask ${\bm{x^t}}$. Then we prepare five 3D tensors all filled with $0$, \ie, $\{\hat {\bm{f}}_i\}_{i=1\dots 5}$. Every tensor has the same height and width with the mask feature tensor, and the same channel number with each component feature tensor. Next, we copy each of the five learned component feature tensors to all-zero tensor $\hat {\bm{f}}_i$ centered at $c_i$ according to the target mask (\eg, mouth-to-mouth, eye-to-eye etc.). After that, we concatenate all 3D and mask feature tensors to produce a fused feature tensor. Finally, we feed the fused feature tensor to the network $G_{m}$ and produce the foreground face image.
\vspace{-1.0em}

\paragraph{Background Fusing Sub-Network.}
To paste the generated foreground faces to the background of the target image, the straightforward approach is to copy the background from the target image $\bm{x^t}$ and combine it with the foreground faces according to the target face mask. However, this causes noticeable boundary artifacts in the final results. There are two possible reasons. First, the background contains neck skin parts, so the unmatched face skin color in the source face $\bm{x^s}$ and the neck skin color in the target image $\bm{x^t}$ cause the artifacts. Second, the segmentation mask for the hair part is not always perfect, so the hair in the background also causes artifacts. 

To solve this problem, we propose using the \emph{background fusing sub-network} to remove the artifacts in fusion. We first use the face parsing network $P_F$ (details in Section~\ref{sec:loss functions}) to get the target face mask $\bm{x^t}$. According to the face mask, we extract the background part from the target image, and then feed the background part to an encoder $E_{b}$ to obtain the output background feature tensor. After that, we concatenate the background feature tensor with the foreground face, and feed it to the generative network $G_{b}$ producing facial result $G(\bm{x^s}, \bm{m^s}, \bm{x^t}, \bm{m^t})$.

\subsection{Loss Functions}
\label{sec:loss functions}
\paragraph{Local Reconstruction.} We use the MSE loss between the input instances and the reconstructed instances to learn the feature embedding of each instance. 
\begin{equation}
\label{eqn:L_local}
\mathcal{L}_{local} = \frac{1}{2}||\bm{x_i^s}- G_{local}^i(E_{local}^i(\bm{x_i^s}))||_2^2,
\vspace{-0.5em}
\end{equation}
where $\bm{x_i^s}$ , $i \in \{0, 1, 2, 3, 4\}$ represents ``left eye", ``right eye", ``mouth", ``skin \& nose", and ``hair" in $\bm{x^s}$. \vspace{-1.0em}

\paragraph{Global Reconstruction.} 
We consider the reconstruction error in training. When the input source images $\bm{x^s}$ is the same as the target image $\bm{x^t}$, the generated result $G(\bm{x^s}, \bm{x^s}, \bm{x^t}, \bm{m^t})$ should be the same as $\bm{x^s}$. Based on the constraint, the reconstruction loss can be measured by:
\begin{equation}
\label{eqn:L_global}
\mathcal{L}_{global} = \frac{1}{2}|| G(\bm{x^s}, \bm{m^s}, \bm{x^t}, \bm{m^t}) - \bm{x^s}||_2^2
\vspace{-1.0em}
\end{equation} \vspace{-1.0em}

\paragraph{Adversarial Loss.}
\label{sec:Adversarial Loss}
To produce realistic results, we add discriminator networks $D$ after the framework. Similar to GAN, the overall framework $G$ plays a minimax game with discriminator network $D$. Since a simple discriminator network $D$ is not suitable for face image synthesis with resolution $256 \times 256$. Following the method in pix2pixHD~\cite{wang2017high}, we also use multi-scale discriminators. We use 2 discriminators that have an identity network structure but operate at different image scales. we downsample the real and generated samples by a factor of 2 using the average pooling layer. Moreover, the generated samples should be conditioned on the target mask $\bm{m^t}$. So the loss function for the discriminators $D_{i}, i \in {1,2}$ is:\vspace{-0.5em}
\begin{small}
\begin{equation}
\label{eqn:L_D}
\begin{aligned}
\mathcal{L}_{D_i}  = &-\mathbb{E}_{\bm{x^t} \thicksim P_{r}}[\mathrm{log} D_i(\bm{x^t}, \bm{m^t})] \\
&- \mathbb{E}_{\bm{x^t},\bm{x^s} \thicksim P_{r}}[\mathrm{log} (1 - D_i(G(\bm{x^s}, \bm{m^s}, \bm{x^t}, \bm{m^t}), \bm{m^t})],
\end{aligned}
\end{equation}
\end{small}
and the loss function for the framework $G$ is:
\vspace{-0.5em}
\begin{small}
\begin{equation}
\label{eqn:L_GD}
\begin{aligned}
\mathcal{L}_{sigmoid}  = - \mathbb{E}_{\bm{x^t},\bm{x^s} \thicksim P_{r}}[\mathrm{log} (D_i(G(\bm{x^s}, \bm{m^s}, \bm{x^t}, \bm{m^t}), \bm{m^t})].
\end{aligned}
\end{equation}
\end{small}
\vspace{-0.8em}

The original loss function for G may cause unstable gradient problems. Inspired by ~\cite{bao2017cvae,wang2017high}, we also use a pairwise feature matching objective for the generator. To generate realistic face images quality,  we match the features of the network $D$ between real and fake images. Let $\bm{f_{D_{i}}}(\bm{x}, \bm{m^t})$ denote features on an intermediate layer of the discriminator, then the pairwise feature matching loss is the Euclidean distance between the feature representations, {\em i.e.},\vspace{-1.5em}

\begin{small}
\begin{equation}
\label{eqn:L_GD_stat}
\mathcal{L}_{FM} =  \frac{1}{2}||\bm{f_{D_{i}}}(G(\bm{x^s}, \bm{m^s}, \bm{x^t}, \bm{m^t}), \bm{m^t}) - \bm{f_{D_{i}}}(\bm{x^t}, \bm{m^t})||_2^2,
\end{equation}
\end{small}
where we use the last output layer of network $D_{i}$ as the feature $\bm{f_{D_{i}}}$ for our experiments.

The overall loss from the discriminator networks to the framework $G$ is:
\vspace{-0.5em}
\begin{equation}
\vspace{-0.5em}
\label{eqn:L_GD}
\mathcal{L}_{GD}  =\mathcal{L}_{sigmoid} + \lambda_{FM} \mathcal{L}_{FM}.
\end{equation}
where $\lambda_{FM}$  controls the importance of the two terms.
\vspace{-1.0em}

\paragraph{Face Parsing Loss.}
\label{sec:Face Parsing Loss}
In order to generate mask equivariant facial images, we need to make the generated samples have the same mask as the target mask, and a face paring network $P_F$ to constrain the generated faces, following previous methods~\cite{long2015fully,ronneberger2015u}. We pretrain the face parsing network $P_F$ with a U-Net network structure on the Helen Face Dataset~\cite{smith2013exemplar}. The loss function $\mathcal{L}_P$  for network $P_F$ is the pixel-wise cross entropy loss, This loss examines each pixel individually, comparing the class predictions (depth-wise pixel vector) to our one-hot encoded target vector $p_{i,j}$:\vspace{-1.0em}

\begin{equation}
\label{eqn:L_P}
\mathcal{L}_P =  -\mathbb{E}_{\bm{x} \thicksim P_{r}}[\sum_{i,j} \log{P(p_{i,j}|P_F(\bm{x})_{i,j})}].
\vspace{-0.5em}
\end{equation}
Here, the $(i,j)$ indicates the location of the pixel.

After we get the pretrained network $P_F$. we use $P_F$ to encourage the generated samples to have the same mask with the target mask, so we use the following loss function for the generative network: 
\vspace{-0.5em}
\begin{small}
\begin{equation}
\label{eqn:L_GP}
\mathcal{L}_{GP} =  -\mathbb{E}_{\bm{x} \thicksim P_{r}}[\sum_{i,j} \log{P(\bm{m^t}_{i,j}|P_F(G(\bm{x^s}, \bm{m^s}, \bm{x^t}, \bm{m^t}))_{i,j})}],
\end{equation}
\end{small}
\vspace{-0.2em}
where $\bm{m^t}_{i,j}$ is the ground truth label of $\bm{x^t}$ located at $(i, j)$. $P_F(G(\bm{x^s}, \bm{m^s}, \bm{x^t}, \bm{m^t}))_{i,j}$ is the predict pixel located at $(i, j)$.
\vspace{-1.0em}

\paragraph{Overall Loss Functions.}
The final loss for $G$ is a sum of the above losses in Equation~\ref{eqn:L_local},~\ref{eqn:L_global},~\ref{eqn:L_GD},~\ref{eqn:L_GP}.
\begin{equation}
\mathcal{L}_{G} = \lambda_{local} \mathcal{L}_{local} + \lambda_{global}\mathcal{L}_{global} + \lambda_{GD} \mathcal{L}_{GD}+\lambda_{GP} \mathcal{L}_{GP},
\end{equation}
where  $\lambda_{local}$, $\lambda_{global}$, $\lambda_{GD}$, and $\lambda_{GP}$
are the trade-offs balancing different losses. In our experiments, $\lambda_{local}$, $\lambda_{global}$, $\lambda_{GD}$, and $\lambda_{GP}$ are set as $\{10, 1, 1, 1\}$ respectively.

\subsection{Training Strategy}
\label{sec:Training Strategy}
During training, the input masks $\bm{m^s}$, $\bm{m^t}$ always use the parsing results of source image $\bm{x^s}$ and target image $\bm{x^t}$. We consider two situations in training: 1) $\bm{x^s}$ and $\bm{x^t}$ are the same, which is called paired data, 2) $\bm{x^s}$ and $\bm{x^t}$ are different, which is called unpaired data. Inspired by ~\cite{bao2018towards} , we incorporate these settings into the training stage, and employ a $(1+1)$ strategy, one step for paired data training and the other step for unpaired data training. However, the training loss functions for these two settings should be different. For paired data, we use all losses in $\mathcal{L}_G$, but for unpaired data, we set $\lambda_{global}$ and $\lambda_{FM}$ to zero in $\mathcal{L}_G$. 

\begin{figure}[t]
\centering
\footnotesize
\includegraphics[width=1\linewidth]{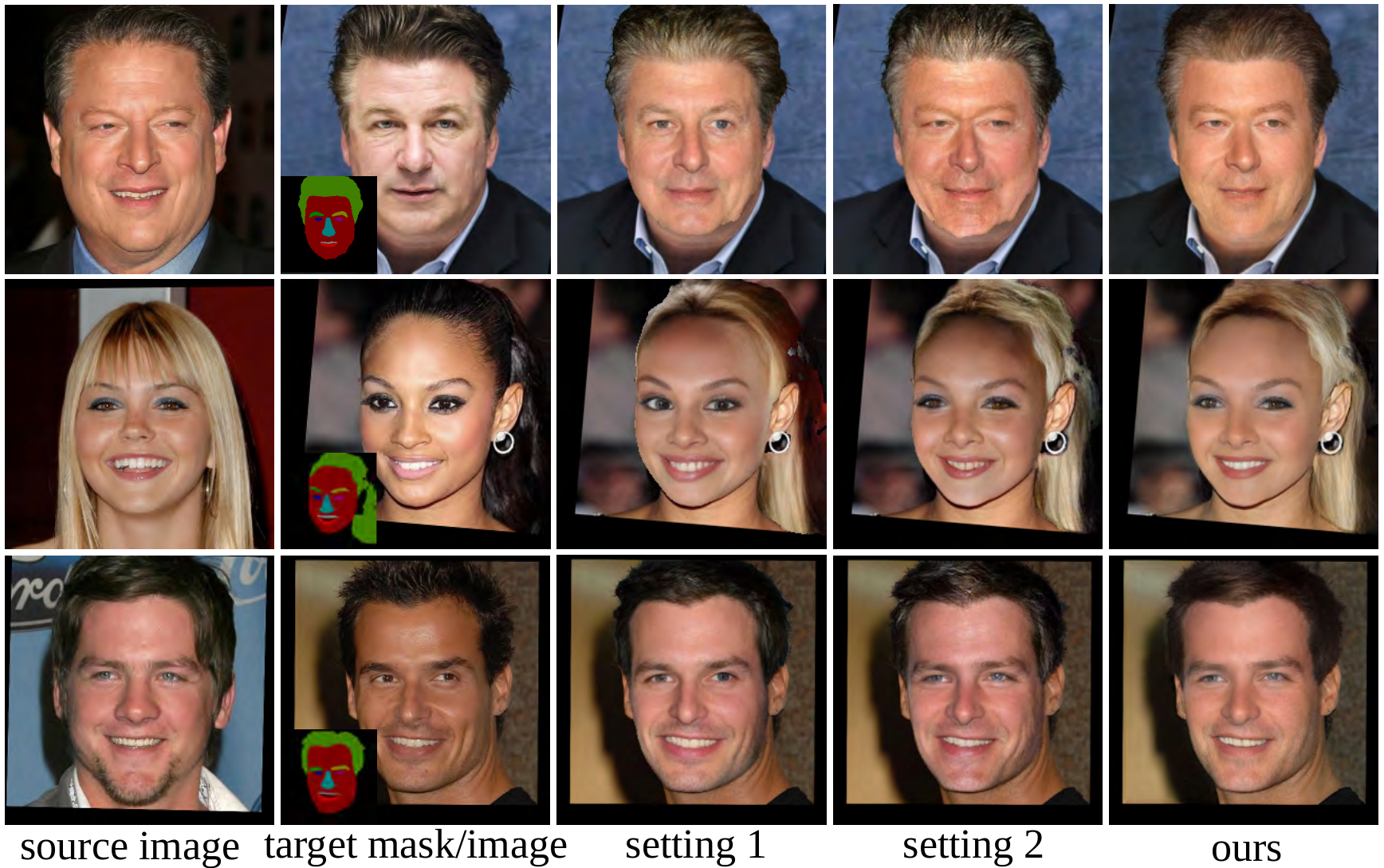}
	\vspace{-1.5em}
    \caption{Visual comparison of our proposed framework and its variants.}
\label{fig:ablation1}
\vspace{-2.0em}
\end{figure}

\section{Experiments}
In this section, we validate the effectiveness of the proposed method. We evaluate our model on the Helen Dataset\cite{smith2013exemplar}. The Helen Dataset contains $2,330$ face images ($2,000$ for training and $330$ for testing) with the pixel-level mask label annotated. But $2,000$ facial images have limited diversity, so we first use these $2,000$ face images to train a face parsing network, and use the parsing network to get semantic masks for an additional $20,000$ face images from VGGFace2~\cite{Cao18}. We use a total of $22,000$ face images for training during experiments . For all training faces, we first detect the facial region with the JDA face detector ~\cite{chen2014joint}, and then locate five facial landmarks (two eyes, nose tip and two mouth corners). After that, we use similarity transformation based on the facial landmarks to align faces to a canonical position. Finally, we crop a $256 \times 256$ facial region to do the experiments.

In our experiments, the input size of five instances (left eye, right eye, mouth, skin, and hair) are decided by the max size of each component. Specially, we use $48 \times 32$, $48 \times 32$, $144 \times 80$, $256 \times 256$, $256 \times 256$ for left eye, right eye, mouth, skin, and hair in our experiments. For network details of $E_{local}$, $G_{local}$, $E_{m}$, $G_{m}$, $E_{b}$, and $G_{b}$ and training settings, please refer to the supplementary material.

\begin{figure}[t]
\centering
\footnotesize
\includegraphics[width=1\linewidth]{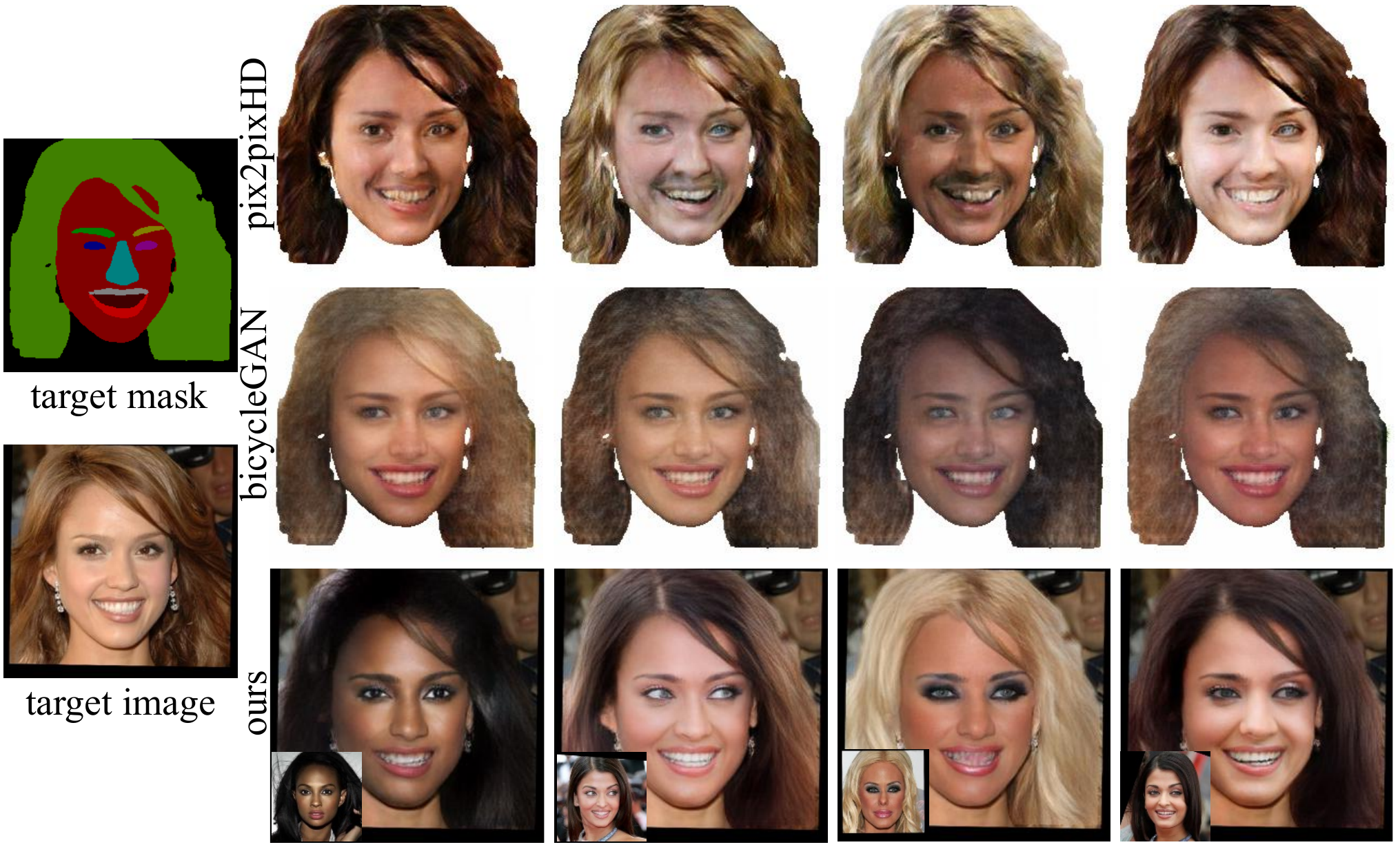}
	\vspace{-1.5em}
    \caption{Comparison of mask to face synthesis results with pix2pixHD~\cite{wang2017high} and BicycleGAN\cite{zhu2017toward}, The target mask and the target face image are on the left side. The first and second rows are the generated results from pix2pixHD and BicyleGAN respectively. The diversity of the generated samples are mainly from the skin color or illumination. The third row is the generated results from our methods. We can generate more realistic and diverse facial images.}
\label{fig:mask2image}
\vspace{-1.5em}
\end{figure}

\begin{figure*}[t]
\centering
\footnotesize
\includegraphics[width=1\linewidth]{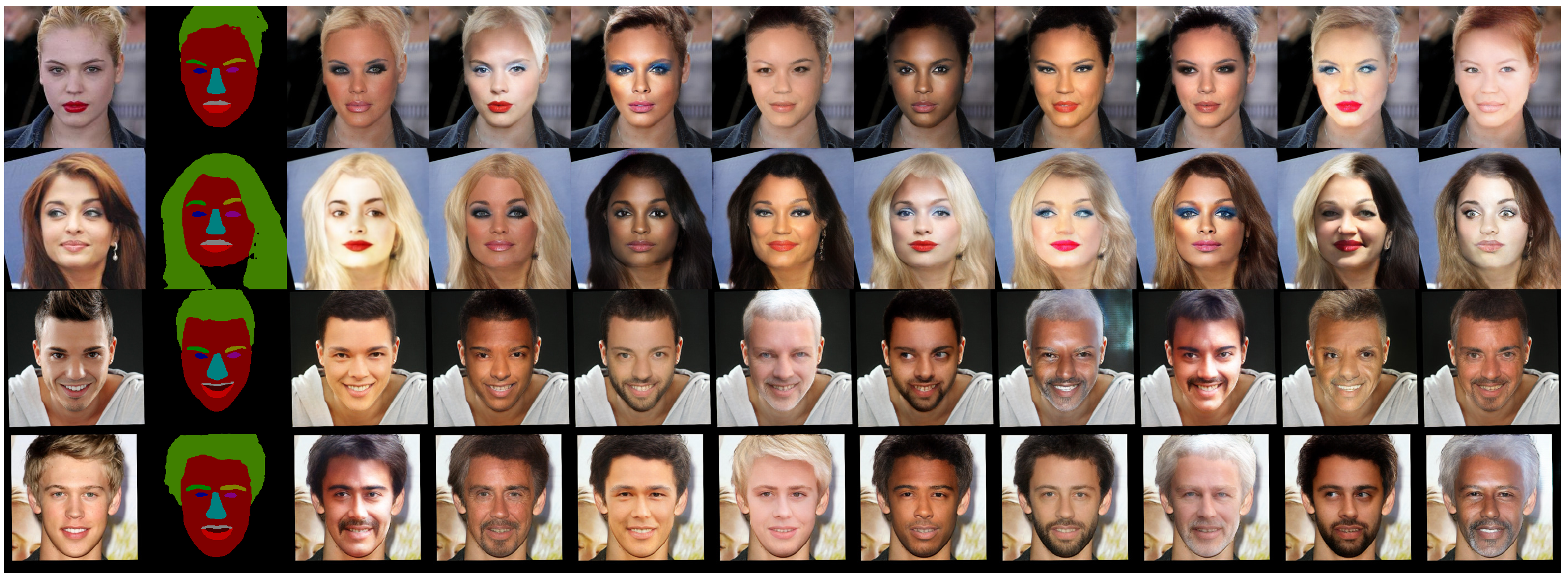}
	\vspace{-2.0em}
    \caption{Our framework can synthesize realistic, diverse and mask equivariant faces from one target mask.}
\label{fig:mask2face_results}
\vspace{-1.5em}
\end{figure*}

\subsection{Analysis of the Proposed Framework}
\label{sec:analysis}
Our framework is designed to solve three problems --diversity, quality, and controllability in mask-guided facial synthesis. To validate this, we perform a step by step ablation study to understand how our proposed framework helps solve these three problems.


We perform three gradually changed settings to validate our framework: 1) We train a framework using a global auto-encoder to learn the global embedding of the source image, then we concatenate the global embedding with the target mask to generate the foreground face image with losses $\mathcal{L}_{global}$, $\mathcal{L}_{GP}$, and $\mathcal{L}_{GD}$. We then crop the background from the target image and directly paste it to the generated foreground face to get the result. 2) We train another framework using a \emph{local embedding sub-network} to learn the embedding of each component of the source image, then we concatenate the local embedding with the target mask to generate the foreground face. After that, we get the background using the same method as 1). 3) We train our framework taking full advantage of \emph{local embedding sub-network}, \emph{mask-guided generative sub-network}, and \emph{background fusing sub-network} to get the final results.

Figure~\ref{fig:ablation1} presents qualitative results for the above three settings. Comparing settings 2 and 1, we see that using a \emph{local embedding sub-network} helps the generated results to keep the details (\eg eye's size, skin color, hair color) from the source images. This enables the controllability of our framework to control each component of the generated face. By feeding different components to the \emph{local embedding sub-network}, we can generate diverse results, which shows our framework handles the diversity problem. Comparing these two variant settings with our method, a background copied directly from the target image causes noticeable boundary artifacts. In our framework, the \emph{background fusing sub-network} helps to remove the artifacts and generate more realistic faces, proving that our framework can generate high quality faces.

To quantitatively evaluate each setting, we generate $5,000$ facial images for each setting, and calculate the FID~\cite{heusel2017gans} between the generated faces and the training faces. In Table~\ref{table:FID_score}, we report the FID for each setting. The \emph{local embedding sub-network} and \emph{background fusing sub-network} help improve the quality of generated samples. Meanwhile, the low FID score indicates that our model can generate high-quality faces from masks.

To validate whether our face parsing loss helps keep the mask of the generated samples, we conduct an experiment to validate this. We train another framework without using the face parsing loss. Then we generate $5,000$ samples from this framework. Next, we use another set of face parsing networks to get the average per-pixel accuracy with the target mask as ground truth for all generated faces. Table~\ref{table:FCN_score} reports the results, showing that the face parsing loss helps to preserve the mask of the generated faces.

\begin{table}[t]
\centering
\caption{Quantitative comparison of our framework and its variants, setting 1,2  are defined in Section~\ref{sec:analysis}.}
\vspace{-0.5em}
\label{table:FID_score}
\small
\begin{tabular}{|c|c|c|c|c|}
 \hline
 Setting  &  1  & 2   & ours \\
 \hline
 FID & 11.02  & 11.26  & 8.92 \\
 \hline
\end{tabular}
\vspace{-1.5em}
\end{table}

\begin{table}[t]
\centering
\caption{Comparison of face parsing results with and without using face parsing networks.}\label{table:FCN_score}
\vspace{-0.5em}
\small
\begin{tabular}{|c|c|c|}
 \hline
 Method & avg. per-pixel accuracy \\
 \hline
w/o face parsing networks & 0.946 \\
 \hline
w face parsing networks & 0.977 \\
 \hline
\end{tabular}
\vspace{-1.5em}
\end{table}

\subsection{Mask-to-Face Synthesis}
This section presents the results of mask to face synthesis. The goal of mask to face synthesis is to generate realistic, diverse and mask equivariant facial images from a given target mask. To demonstrate that our framework has the ability to generate realistic and diverse faces from an input mask, we choose some masks and randomly choose some facial images as the source image and synthesize the facial images. 

Figure~\ref{fig:mask2face_results} presents the face synthesis results from the input masks. The first column is the target masks, and the face images on the right side are the generated face images conditioned on the target masks. We observe that the generated face images are photo-realistic. Meanwhile, the generated facial images perform well in terms of diversity, such as in skin color, hair color, eye makeup, and even beard. Furthermore, the generated facial images also maintain the mask.

Previous methods also try to do the mask-to-face synthesis, BicycleGAN~\cite{zhu2017toward} is an image-to-image translation model which can generate continuous and multimodal output distributions. pix2pixHD allows high resolution image-to-image translation. In Figure ~\ref{fig:mask2image}, we show the qualitative comparison results. We observe that the generated samples by BicycleGAN and pix2pixHD show a limited diversity, and the diversity lies in the skin color or illumination. The reason for this is that they use a global encoder, so the generated samples cannot leverage the diversity of components in the faces. In contrast, the generated results by our methods look realistic, clear and diverse. Our model
is also able to keep the mask information. It shows the strength of the proposed framework.


\subsection{Face Editing and Face Swap+}
Another important application for our framework is facial editing. With the changeable target mask and source image, we can edit on the generated face images by editing the target mask or replacing the facial component from the source image. We conduct two experiments: 1) changing the target mask to change the appearance of the generated faces; 2) replacing the facial component of the target image with new component from other's faces to explicitly change the corresponding component of the generated faces.

\begin{figure}[t]
\centering
 \includegraphics[width=0.95\columnwidth]{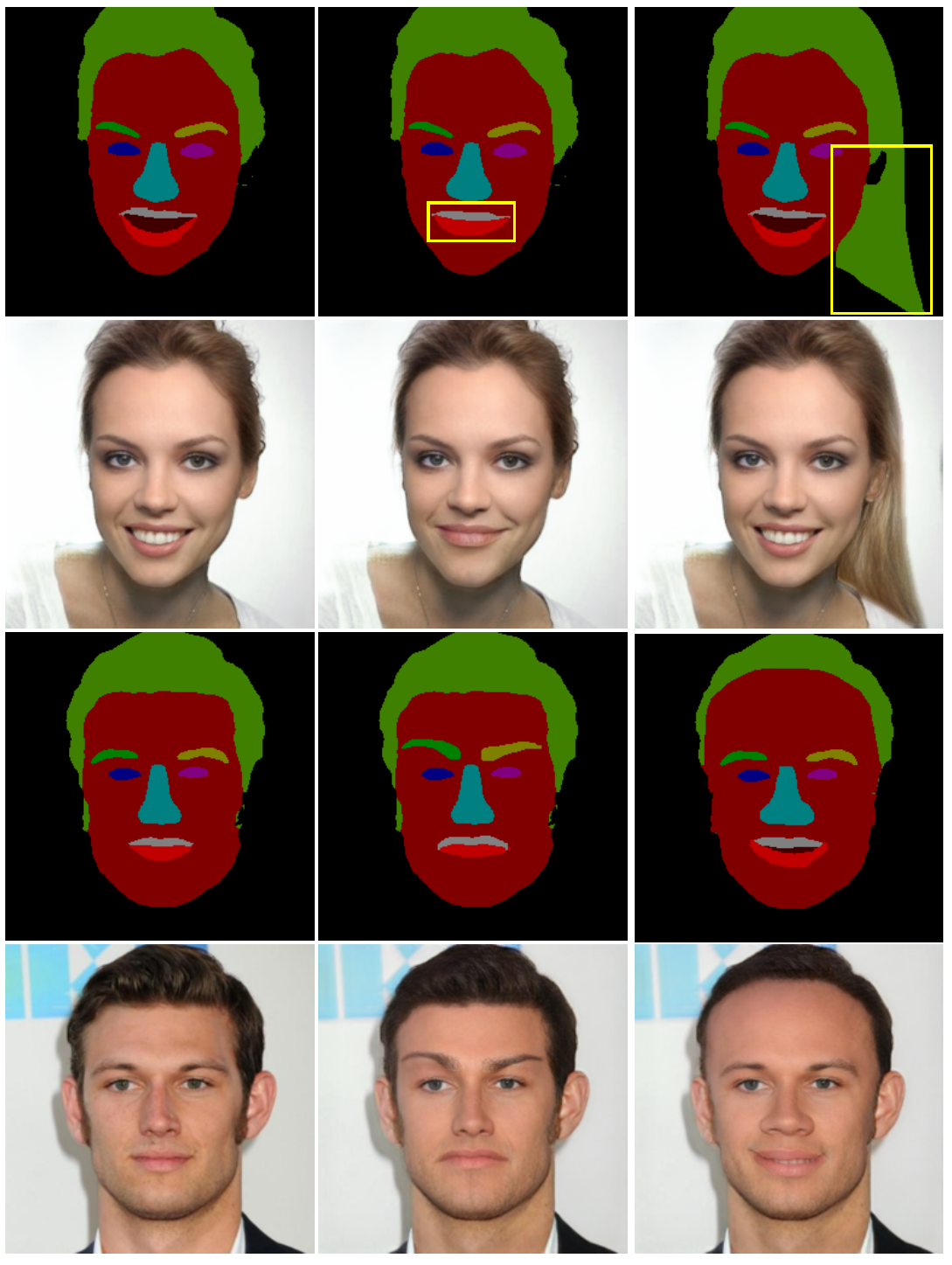}
 \footnotesize
 	\vspace{-0.5em}
    \caption{Our framework allows users to change mask labels locally to explicitly manipulate facial components, like making short hair become longer, and even changing the emotion.}
\label{fig:edit_mask}\vspace{-0.7em}
\vspace{-1.5em}
\end{figure}

\begin{figure}[t]
\centering
 \includegraphics[width=0.95\columnwidth]{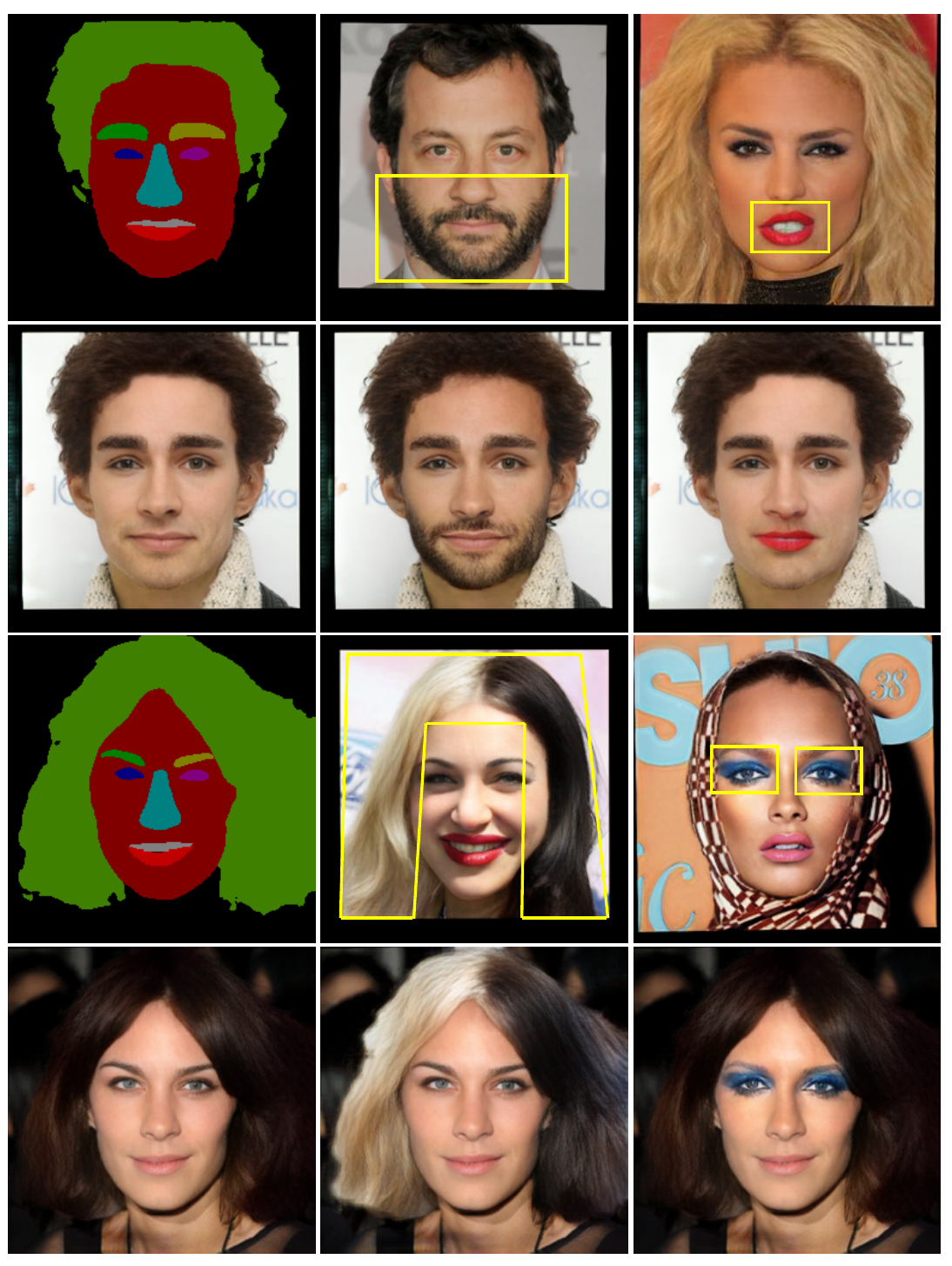}
 \footnotesize
 	\vspace{-1.0em}
    \caption{We can also edit the appearance of instances in the face, {\em e.g.} adding beards and changing the color of mouth, hair, and eye makeup.}
\label{fig:edit_content}\vspace{-0.7em}
\vspace{-0.5em}
\end{figure}

Figure~\ref{fig:edit_mask} shows the generated results of changing the target mask. We replace the label of hair region on forehead with skin label, and we get a vivid face images with no hair on the forehead. Besides that, we can change the shape of the mouth, eyes, eyebrows region in the mask, and get an output facial image even with new emotions. This shows the effectiveness of our model in generating mask equivariant and realistic results.

Figure~\ref{fig:edit_content} shows the results of changing individual parts of the generated face. We can add the beards to the generated face by replacing the target skin part with the skin part of a face with a beard. We also change the mouth color, hair color, and even the eye makeup by changing the local embedding part. This shows the effectiveness of our model in generating local instance equivariant and realistic results. 

\begin{figure}[t]
\centering
 \includegraphics[width=0.95\columnwidth]{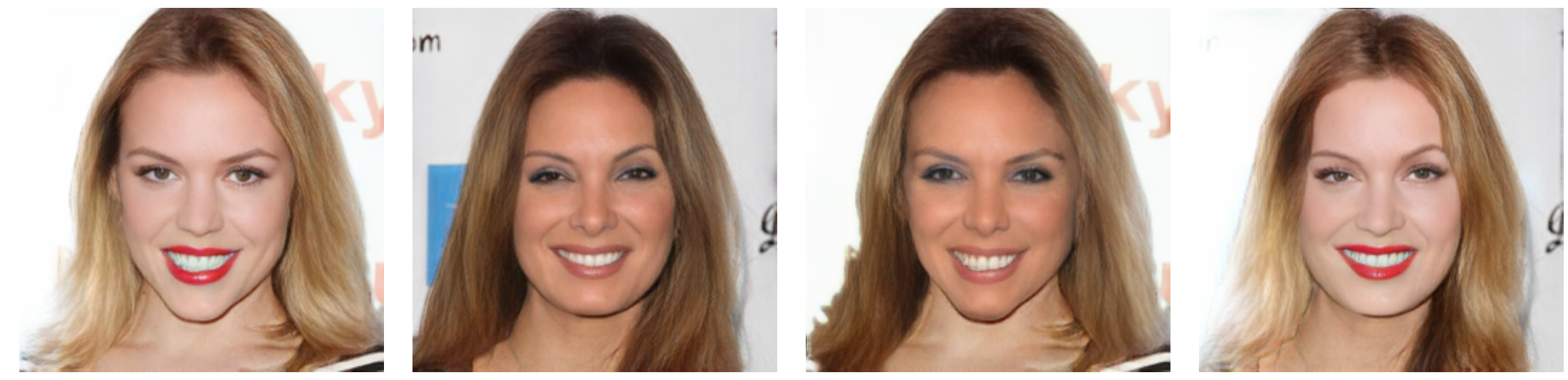}
 \footnotesize
 	\vspace{-0.5em}
    \caption{Our framework can enhance an existing facial swap, called face swap+, to explicitly swap the hair.}
\label{fig:face_swap}\vspace{-0.7em}
\end{figure}

Figure~\ref{fig:face_swap} shows the results of face swap+. Different from traditional face swap algorithm, our framework can 
not only swap the face appearance but also keep the component shape. More importantly, we can explicitly swap the hair part compared to previous methods.

\begin{figure}[t]
\centering
 \includegraphics[width=0.95\columnwidth]{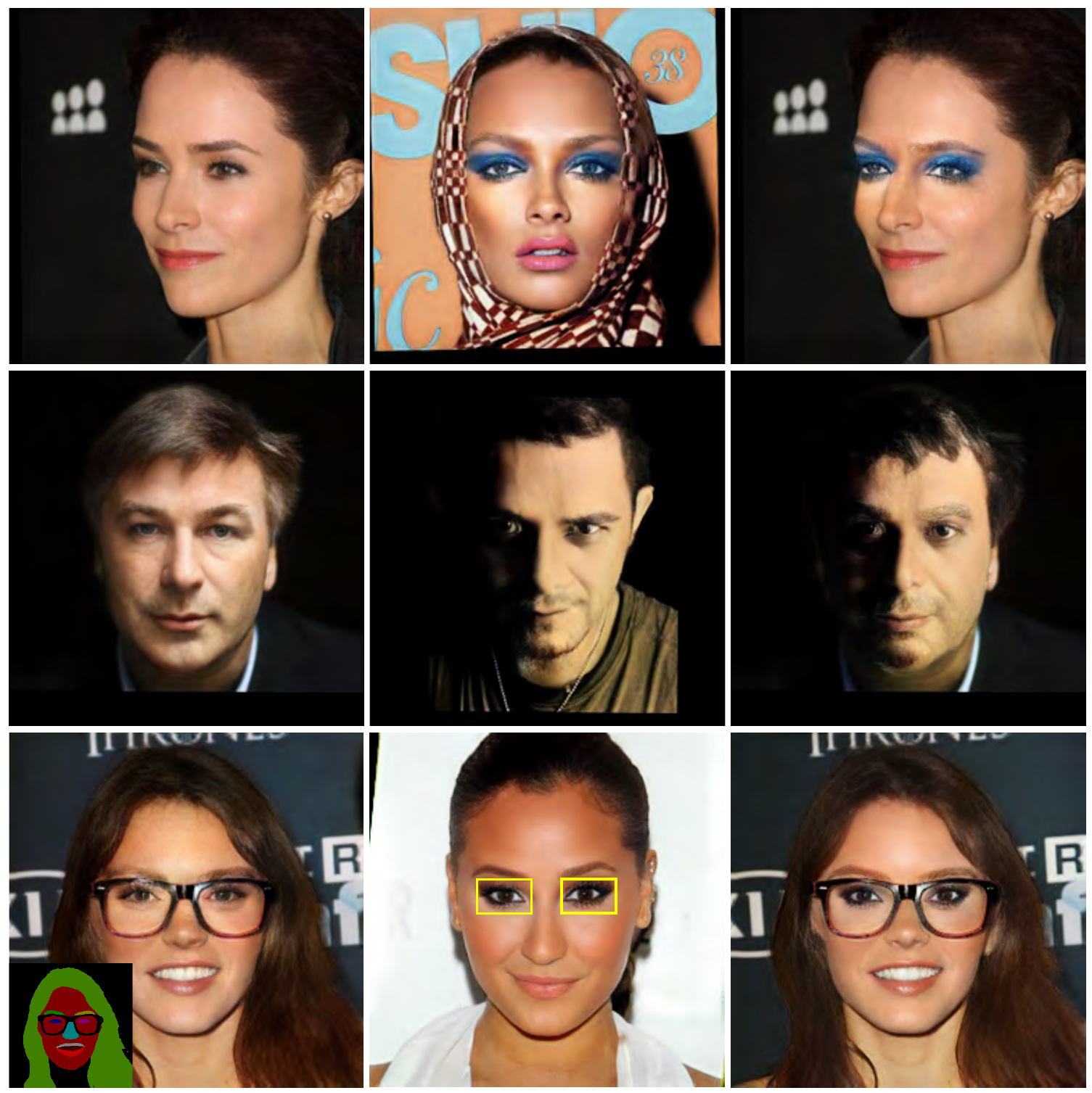}
 \footnotesize
 	\vspace{-0.5em}
    \caption{The generated results with input faces under extreme conditions (large pose, extreme illumination and face with glasses).}
\label{fig:face_extreme}\vspace{-1.0em}
\end{figure}

Furthermore, Figure~\ref{fig:face_extreme} shows more results for input face under extreme conditions. In the first two rows, the input face images have large poses and extreme illumination. Our method gets reasonable and realistic results. Also, if the input face has eye-glasses, we find that the result relies on the segmentation mask. If the glasses are labeled as background, it can be reconstructed by our \emph{background fusion sub-network}, the generated result is shown in the last row.

\subsection{Synthesized Faces for Face Parsing}

We further show that the facial images synthesized from our framework can benefit the training the face parsing model. We use the $2,000$ masks of the trainings images of Helen Face Dataset as the target mask, and randomly choose face images from CelebA Dataset ~\cite{liu2015faceattributes} as the source image to generate face images, then we remove these facial images generated with different genders. Then we conduct three experiments: 1) we only use the training set of the Helen Face Dataset to train a face parsing network without data augmentation; 2) we use the training set of the Helen Face Dataset to train a face parsing network with a data augmentation strategy (flip, geometry transform, scale, rotation); 3) we add the generated facial images to the Helen Face Dataset training set, using target mask as ground-truth and the same data augmentation strategy.

\begin{table}[t]
\vspace{-1pt}
\centering
\caption{Results of face parsing with added generated facial images.}\label{table:data_aug}
\vspace{-0.5em}
\small
\begin{tabular}{|c|c|}
 \hline
Helen & 0.728  \\
  \hline
Helen (with data augmentation) & 0.863  \\
 \hline
Helen + generated (with data augmentation)  & 0.871  \\
 \hline
\end{tabular}
\vspace{-1.0em}
\end{table}

We use $100$ test images from the Helen Face Dataset for testing. Table ~\ref{table:data_aug} shows the face parsing accuracy in three different settings. With the new generated faces, we get a $0.8\%$ improvement in accuracy compared with no generated face images. This demonstrates that our generative framework has a certain extrapolation ability.
\vspace{-0.5em}

\section{Conclusion}
In this paper, we propose a novel end-to-end framework based on mask-guided conditional GANs, which can synthesize diverse, high-quality, and controllable facial images from given masks. With the changeable input facial mask and source image, our framework allows users to do high-level portrait editing, such as: explicitly editing face components in the generated face and transferring local appearances from one face to the generated face. We can even get a better face paring model by leveraging the synthesized facial data from input masks. Our experiments demonstrate the excellent performance of our framework.

 
{\small
\bibliographystyle{ieee}
\bibliography{egbib}

\begin{thebibliography}{10}\itemsep=-1pt

\bibitem{antipov2017face}
G.~Antipov, M.~Baccouche, and J.-L. Dugelay.
\newblock Face aging with conditional generative adversarial networks.
\newblock In {\em Image Processing (ICIP), 2017 IEEE International Conference
  on}, pages 2089--2093. IEEE, 2017.

\bibitem{arjovsky2017wasserstein}
M.~Arjovsky, S.~Chintala, and L.~Bottou.
\newblock Wasserstein gan.
\newblock {\em arXiv preprint arXiv:1701.07875}, 2017.

\bibitem{bao2017cvae}
J.~Bao, D.~Chen, F.~Wen, H.~Li, and G.~Hua.
\newblock Cvae-gan: Fine-grained image generation through asymmetric training.
\newblock In {\em 2017 IEEE International Conference on Computer Vision
  (ICCV)}, pages 2764--2773. IEEE, 2017.

\bibitem{bao2018towards}
J.~Bao, D.~Chen, F.~Wen, H.~Li, and G.~Hua.
\newblock Towards open-set identity preserving face synthesis.
\newblock In {\em Proceedings of the IEEE Conference on Computer Vision and
  Pattern Recognition}, pages 6713--6722, 2018.

\bibitem{bitouk2008face}
D.~Bitouk, N.~Kumar, S.~Dhillon, P.~Belhumeur, and S.~K. Nayar.
\newblock Face swapping: automatically replacing faces in photographs.
\newblock In {\em ACM Transactions on Graphics (TOG)}, volume~27, page~39. ACM,
  2008.

\bibitem{busto2010instant}
P.~P. Busto, C.~Eisenacher, S.~Lefebvre, M.~Stamminger, et~al.
\newblock Instant texture synthesis by numbers.
\newblock In {\em VMV}, pages 81--85, 2010.

\bibitem{Cao18}
Q.~Cao, L.~Shen, W.~Xie, O.~M. Parkhi, and A.~Zisserman.
\newblock Vggface2: A dataset for recognising faces across pose and age.
\newblock In {\em International Conference on Automatic Face and Gesture
  Recognition}, 2018.

\bibitem{champandard2016semantic}
A.~J. Champandard.
\newblock Semantic style transfer and turning two-bit doodles into fine
  artworks.
\newblock {\em arXiv preprint arXiv:1603.01768}, 2016.

\bibitem{chang2018pairedcyclegan}
H.~Chang, J.~Lu, F.~Yu, and A.~Finkelstein.
\newblock {PairedCycleGAN}: Asymmetric style transfer for applying and removing
  makeup.
\newblock In {\em 2018 IEEE Conference on Computer Vision and Pattern
  Recognition (CVPR)}, 2018.

\bibitem{chen2014joint}
D.~Chen, S.~Ren, Y.~Wei, X.~Cao, and J.~Sun.
\newblock Joint cascade face detection and alignment.
\newblock In {\em European Conference on Computer Vision}, pages 109--122.
  Springer, 2014.

\bibitem{chen2017photographic}
Q.~Chen and V.~Koltun.
\newblock Photographic image synthesis with cascaded refinement networks.
\newblock In {\em IEEE International Conference on Computer Vision (ICCV)},
  volume~1, page~3, 2017.

\bibitem{choi2017stargan}
Y.~Choi, M.~Choi, M.~Kim, J.-W. Ha, S.~Kim, and J.~Choo.
\newblock Stargan: Unified generative adversarial networks for multi-domain
  image-to-image translation.
\newblock {\em arXiv preprint}, 1711, 2017.

\bibitem{dolhansky2018eye}
B.~Dolhansky and C.~Canton~Ferrer.
\newblock Eye in-painting with exemplar generative adversarial networks.
\newblock In {\em Proceedings of the IEEE Conference on Computer Vision and
  Pattern Recognition}, pages 7902--7911, 2018.

\bibitem{fivser2017example}
J.~Fi{\v{s}}er, O.~Jamri{\v{s}}ka, D.~Simons, E.~Shechtman, J.~Lu, P.~Asente,
  M.~Luk{\'a}{\v{c}}, and D.~S{\`y}kora.
\newblock Example-based synthesis of stylized facial animations.
\newblock {\em ACM Transactions on Graphics (TOG)}, 36(4):155, 2017.

\bibitem{gatys2016image}
L.~A. Gatys, A.~S. Ecker, and M.~Bethge.
\newblock Image style transfer using convolutional neural networks.
\newblock In {\em 2016 IEEE Conference on Computer Vision and Pattern
  Recognition (CVPR)}, pages 2414--2423. IEEE, 2016.

\bibitem{goodfellow2014generative}
I.~Goodfellow, J.~Pouget-Abadie, M.~Mirza, B.~Xu, D.~Warde-Farley, S.~Ozair,
  A.~Courville, and Y.~Bengio.
\newblock Generative adversarial nets.
\newblock In {\em Advances in neural information processing systems}, pages
  2672--2680, 2014.

\bibitem{gu2018arbitrary}
S.~Gu, C.~Chen, J.~Liao, and L.~Yuan.
\newblock Arbitrary style transfer with deep feature reshuffle.
\newblock In {\em Proceedings of the IEEE Conference on Computer Vision and
  Pattern Recognition}, pages 8222--8231, 2018.

\bibitem{hays2007scene}
J.~Hays and A.~A. Efros.
\newblock Scene completion using millions of photographs.
\newblock In {\em ACM Transactions on Graphics (TOG)}, volume~26, page~4. ACM,
  2007.

\bibitem{he2017neural}
M.~He, J.~Liao, L.~Yuan, and P.~V. Sander.
\newblock Neural color transfer between images.
\newblock {\em arXiv preprint arXiv:1710.00756}, 2017.

\bibitem{heusel2017gans}
M.~Heusel, H.~Ramsauer, T.~Unterthiner, B.~Nessler, and S.~Hochreiter.
\newblock Gans trained by a two time-scale update rule converge to a local nash
  equilibrium.
\newblock In {\em Advances in Neural Information Processing Systems}, pages
  6626--6637, 2017.

\bibitem{huang2017beyond}
R.~Huang, S.~Zhang, T.~Li, R.~He, et~al.
\newblock Beyond face rotation: Global and local perception gan for
  photorealistic and identity preserving frontal view synthesis.
\newblock {\em arXiv preprint arXiv:1704.04086}, 2017.

\bibitem{huang2018multimodal}
X.~Huang, M.-Y. Liu, S.~Belongie, and J.~Kautz.
\newblock Multimodal unsupervised image-to-image translation.
\newblock {\em arXiv preprint arXiv:1804.04732}, 2018.

\bibitem{iizuka2017globally}
S.~Iizuka, E.~Simo-Serra, and H.~Ishikawa.
\newblock Globally and locally consistent image completion.
\newblock {\em ACM Transactions on Graphics (TOG)}, 36(4):107, 2017.

\bibitem{isola2017image}
P.~Isola, J.-Y. Zhu, T.~Zhou, and A.~A. Efros.
\newblock Image-to-image translation with conditional adversarial networks.
\newblock {\em arXiv preprint}, 2017.

\bibitem{karras2017progressive}
T.~Karras, T.~Aila, S.~Laine, and J.~Lehtinen.
\newblock Progressive growing of gans for improved quality, stability, and
  variation.
\newblock {\em arXiv preprint arXiv:1710.10196}, 2017.

\bibitem{korshunova2017fast}
I.~Korshunova, W.~Shi, J.~Dambre, and L.~Theis.
\newblock Fast face-swap using convolutional neural networks.
\newblock In {\em The IEEE International Conference on Computer Vision}, pages
  3697--3705, 2017.

\bibitem{lalonde2007photo}
J.-F. Lalonde, D.~Hoiem, A.~A. Efros, C.~Rother, J.~Winn, and A.~Criminisi.
\newblock Photo clip art.
\newblock {\em ACM transactions on graphics (TOG)}, 26(3):3, 2007.

\bibitem{ledig2017photo}
C.~Ledig, L.~Theis, F.~Husz{\'a}r, J.~Caballero, A.~Cunningham, A.~Acosta,
  A.~P. Aitken, A.~Tejani, J.~Totz, Z.~Wang, et~al.
\newblock Photo-realistic single image super-resolution using a generative
  adversarial network.
\newblock In {\em CVPR}, volume~2, page~4, 2017.

\bibitem{li2016convolutional}
M.~Li, W.~Zuo, and D.~Zhang.
\newblock Convolutional network for attribute-driven and identity-preserving
  human face generation.
\newblock {\em arXiv preprint arXiv:1608.06434}, 2016.

\bibitem{liao2017visual}
J.~Liao, Y.~Yao, L.~Yuan, G.~Hua, and S.~B. Kang.
\newblock Visual attribute transfer through deep image analogy.
\newblock {\em arXiv preprint arXiv:1705.01088}, 2017.

\bibitem{liu2015faceattributes}
Z.~Liu, P.~Luo, X.~Wang, and X.~Tang.
\newblock Deep learning face attributes in the wild.
\newblock In {\em Proceedings of International Conference on Computer Vision
  (ICCV)}, 2015.

\bibitem{long2015fully}
J.~Long, E.~Shelhamer, and T.~Darrell.
\newblock Fully convolutional networks for semantic segmentation.
\newblock In {\em Proceedings of the IEEE conference on computer vision and
  pattern recognition}, pages 3431--3440, 2015.

\bibitem{mirza2014conditional}
M.~Mirza and S.~Osindero.
\newblock Conditional generative adversarial nets.
\newblock {\em arXiv preprint arXiv:1411.1784}, 2014.

\bibitem{portenier2018faceshop}
T.~Portenier, Q.~Hu, A.~Szabo, S.~Bigdeli, P.~Favaro, and M.~Zwicker.
\newblock Faceshop: Deep sketch-based face image editing.
\newblock {\em arXiv preprint arXiv:1804.08972}, 2018.

\bibitem{pumarola2018ganimation}
A.~Pumarola, A.~Agudo, A.~M. Martinez, A.~Sanfeliu, and F.~Moreno-Noguer.
\newblock Ganimation: Anatomically-aware facial animation from a single image.
\newblock In {\em Proceedings of the European Conference on Computer Vision
  (ECCV)}, pages 818--833, 2018.

\bibitem{qi2017loss}
G.-J. Qi.
\newblock Loss-sensitive generative adversarial networks on lipschitz
  densities.
\newblock {\em arXiv preprint arXiv:1701.06264}, 2017.

\bibitem{qi2018semi}
X.~Qi, Q.~Chen, J.~Jia, and V.~Koltun.
\newblock Semi-parametric image synthesis.
\newblock In {\em Proceedings of the IEEE Conference on Computer Vision and
  Pattern Recognition}, pages 8808--8816, 2018.

\bibitem{ronneberger2015u}
O.~Ronneberger, P.~Fischer, and T.~Brox.
\newblock U-net: Convolutional networks for biomedical image segmentation.
\newblock In {\em International Conference on Medical image computing and
  computer-assisted intervention}, pages 234--241. Springer, 2015.

\bibitem{sangkloy2017scribbler}
P.~Sangkloy, J.~Lu, C.~Fang, F.~Yu, and J.~Hays.
\newblock Scribbler: Controlling deep image synthesis with sketch and color.
\newblock In {\em IEEE Conference on Computer Vision and Pattern Recognition
  (CVPR)}, volume~2, 2017.

\bibitem{shih2014style}
Y.~Shih, S.~Paris, C.~Barnes, W.~T. Freeman, and F.~Durand.
\newblock Style transfer for headshot portraits.
\newblock {\em ACM Transactions on Graphics (TOG)}, 33(4):148, 2014.

\bibitem{shu2017eyeopener}
Z.~Shu, E.~Shechtman, D.~Samaras, and S.~Hadap.
\newblock Eyeopener: Editing eyes in the wild.
\newblock {\em ACM Transactions on Graphics (TOG)}, 36(1):1, 2017.

\bibitem{shu2017neural}
Z.~Shu, E.~Yumer, S.~Hadap, K.~Sunkavalli, E.~Shechtman, and D.~Samaras.
\newblock Neural face editing with intrinsic image disentangling.
\newblock In {\em Proceedings of the IEEE Conference on Computer Vision and
  Pattern Recognition}, pages 5541--5550, 2017.

\bibitem{smith2013exemplar}
B.~M. Smith, L.~Zhang, J.~Brandt, Z.~Lin, and J.~Yang.
\newblock Exemplar-based face parsing.
\newblock In {\em Proceedings of the IEEE Conference on Computer Vision and
  Pattern Recognition}, pages 3484--3491, 2013.

\bibitem{tran2017disentangled}
L.~Tran, X.~Yin, and X.~Liu.
\newblock Disentangled representation learning gan for pose-invariant face
  recognition.
\newblock In {\em CVPR}, volume~3, page~7, 2017.

\bibitem{wang2017high}
T.-C. Wang, M.-Y. Liu, J.-Y. Zhu, A.~Tao, J.~Kautz, and B.~Catanzaro.
\newblock High-resolution image synthesis and semantic manipulation with
  conditional gans.
\newblock {\em arXiv preprint arXiv:1711.11585}, 2017.

\bibitem{yang2017learning}
H.~Yang, D.~Huang, Y.~Wang, and A.~K. Jain.
\newblock Learning face age progression: A pyramid architecture of gans.
\newblock {\em arXiv preprint arXiv:1711.10352}, 2017.

\bibitem{zhu2017unpaired}
J.-Y. Zhu, T.~Park, P.~Isola, and A.~A. Efros.
\newblock Unpaired image-to-image translation using cycle-consistent
  adversarial networks.
\newblock {\em arXiv preprint}, 2017.

\bibitem{zhu2017toward}
J.-Y. Zhu, R.~Zhang, D.~Pathak, T.~Darrell, A.~A. Efros, O.~Wang, and
  E.~Shechtman.
\newblock Toward multimodal image-to-image translation.
\newblock In {\em Advances in Neural Information Processing Systems}, pages
  465--476, 2017.

\end{thebibliography}
}

\end{document}